\documentclass[a4paper,conference]{IEEEtran}
\usepackage{booktabs}
\usepackage{multirow}
\usepackage{xcolor}
\usepackage[linesnumbered,ruled,vlined]{algorithm2e}
\SetKwInput{KwData}{Input}
\SetKwInput{KwResult}{Output}

\SetCommentSty{mycommfont}

\usepackage[linesnumbered,ruled,vlined]{algorithm2e}
\usepackage{graphicx}
\usepackage{amsmath,amsthm}
\usepackage{amsfonts,amssymb}
\usepackage{subfigure}
\usepackage{authblk}
\ifCLASSINFOpdf

\else

\fi
\begin{document}

\title{Learning to Aggregate and Refine Noisy Labels for Visual Sentiment Analysis}
\author[]{Wei Zhu}
\author[]{Zihe Zheng}
\author[]{Haitian Zheng}
\author[]{Hanjia Lyu}
\author[]{Jiebo Luo}
\affil[]{University of Rochester}

\maketitle

\begin{abstract}
Visual sentiment analysis has received increasing attention in recent years. However, the dataset's quality is a concern because the sentiment labels are crowd-sourcing, subjective, and prone to mistakes, and poses a severe threat to the data-driven models, especially the deep neural networks. The deep models would generalize poorly on the testing cases when trained to over-fit the training samples with noisy sentiment labels. Inspired by the recent progress on learning with noisy labels, we propose a robust learning method to perform robust visual sentiment analysis. Our method relies on external memory to aggregate and filters noisy labels during training. The memory is composed of the prototypes with corresponding labels, which can be updated online. The learned prototypes and their labels can be regarded as denoising features and labels for the local regions and can guide the training process to prevent the model from overfitting the noisy cases. We establish a benchmark for visual sentiment analysis with label noise using publicly available datasets. The experiment results of the proposed benchmark settings comprehensively show the effectiveness of our method.
\end{abstract}

\IEEEpeerreviewmaketitle

\section{Introduction}
\label{sec:intro}
There is increasing attention in visual sentiment analysis driven by the need for more and more people to share their feelings with images, emojis, and other visual content. With the successes of deep neural networks in conventional computer vision tasks, numerous methods have been proposed to conduct visual sentiment analysis and have shown clear advantages over traditional methods with handcrafted features~\cite{Chen2014DeepSentiBankVS}. 

However, there are several issues when applying deep neural networks to visual sentiment analysis~\cite{xue2020nlwsnet,yang2018weakly}. The labels for visual sentiment analysis are inherently subjective and error-prone since it can be confusing for humans to recognize the sentiment of images~\cite{yang2018weakly}. The frequently mislabeled samples will hinder the performance of deep models and cause the models to generalize poorly on unseen cases. 
Several existing datasets are labeled by first querying the search engine with keywords~\cite{You2016BuildingAL}. Although crowd workers were employed to verify the labels data manually, the resulting labels still contain a fair amount of noise. 
Therefore, it is critical to develop robust methods to handle mislabeled samples explicitly. 

This paper proposes a method called Aggregate and Refine Net (ARNet) to mitigate the problem. ARNet is equipped with an external memory composed of prototypes with corresponding pseudo labels. Basically, the prototypes are optimized to be locality representative of the feature space and thus can work as proxies to aggregate the labels of the samples that are close to them to form prototype labels. With the help of the optimized memory, the proposed method can learn to filter the noisy labels and provide a refined label for each training sample. The key to our method is a strategy to read and write the prototypes and the corresponding labels stored in the memory. More specifically, we consider the input features as queries, prototypes as keys, and prototype labels as values. We read the memory through the attention addressing mechanism. The memory-guided labels for the training samples can be retrieved from the memory and used to refine the noisy labels. For memory updating, we write the memory in an online manner during training and update the differentiable prototypes by gradient descent. Subsequently, the prototype labels are updated in a momentum fashion. The refined labels obtained with the external memory can thus guide us to build a more robust model for visual sentiment analysis. 

The main contributions of our paper are as follows:
\begin{enumerate}
    \item We establish a benchmark for visual sentiment analysis with label noise on publicly available datasets to explicitly handle commonly existing noisy sentiment labels.
    \item We address visual sentiment analysis with label noise using an Aggregate and Refine Network (ARNet). ARNet aggregates and filters the noisy labels with an external memory composed of prototypes with labels. The learned prototypes and their labels can be regarded as denoising features and labels for the local regions.
    \item The proposed ARNet achieves superior performance compared with the state-of-the-art learning-with-noisy-labels methods on the established benchmark for visual sentiment analysis with label noise.
\end{enumerate}

\begin{figure*}
\centering
\includegraphics[width=0.75\textwidth]{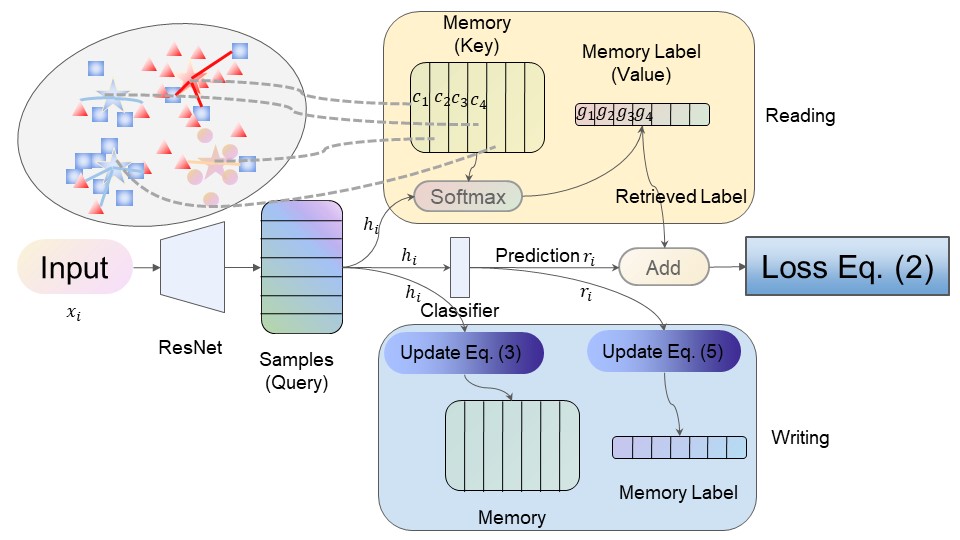}
\caption{Overview of the proposed method. We retrieve the pseudo labels for the training samples according to the external memory through the attention addressing mechanism. The memory comprises the prototypes and the corresponding labels, which are updated during training. We give an example of the latent space during training on the top left side. Stars denote the prototypes stored in the memory, and other shapes are training samples. Different classes are in different colors. The prototypes are updated by a clustering objective (detailed in Sec. \ref{subsec:writing}) to be locality representative. The prototype labels are obtained by aggregating the prediction of nearby samples assigned to the prototypes and enabling the training samples to refine their label by reading and modifying the memory mutually.}
\label{fig:overview}
\end{figure*}
\section{Related Work}
\subsection{Visual Sentiment Analysis}
Early works on visual sentiment analysis are based on handcrafted features. Machajdik and Hanbury~\cite{Machajdik2010AffectiveIC} employed features from art and psychology theories to define image features that are specific to the domain of artworks with emotional expression. Borth \textit{et al.}~\cite{Borth2013SentiBankLO} proposed SentiBank, a detector library containing 1,200 Adjective Noun Pairs (ANP). As a continuation of SentiBank, Chen \textit{et al.}~\cite{Chen2014DeepSentiBankVS} proposed DeepSentiBank, which uses convolutional neural networks (CNNs) to determine which ANPs exist in an image and classify the sentiment of images accordingly. You \textit{et al.} pre-trained a CNN on a half-million-image Flickr dataset~\cite{Borth2013SentiBankLO} and used a progressive fine-tuning process to transfer the knowledge to a smaller dataset~\cite{You2015RobustIS}. Class Activation Maps(CAM)~\cite{Zhou2016LearningDF} is a spatial heat map that shows the discriminative areas of an image in a classification process. CAM-based methods have shown great potential in image classification~\cite{Zhu2017SoftPN}, as well as visual sentiment analysis. She \textit{et al.}~\cite{She2020WSCNetWS} proposed a coupled network (WSCNet) that generates a sentiment CAM to couple with the classification branch. However, little work has been conducted to explicitly handle the noisy and subjective sentiment labels. NLWSNet~\cite{xue2020nlwsnet} was proposed to handle the problem by a non-extreme attention mechanism and a special-class activation map module. However, their work lacked a comparison with the existing methods of learning with noisy labels, and the settings are also not realistic to some extent.

\subsection{Learning with Noisy Label}
With the proliferation of content on the Internet, large-scale datasets can be easily constructed. However,  noisy labels are commonplace. Some previous works show that simply applying supervised methods on weakly-labeled datasets cannot fully exploit the information in such noisy datasets~\cite{Li2020MoProWS}. One of the approaches to perform robust learning on datasets with noisy labels is modifying the cross-entropy loss~\cite{Zhang2018GeneralizedCE,ma2020normalized}. Zhang and Sabuncu~\cite{Zhang2018GeneralizedCE} proposed a generalized cross-entropy loss, which is a generalization of the mean absolute error (MAE) and categorical cross-entropy loss (CCE). Wang \textit{et al.}~\cite{Wang2019SymmetricCE} showed that simple deep neural networks could overfit noisy labels while underfitting hard labels and proposed Symmetric cross-entropy Learning (SL) to address both problems. Knowledge distillation can also be used to refine noisy label~\cite{zhang2020distilling}. Curriculum learning is also a possible method for robust learning~\cite{Bengio2009CurriculumL,huang2020self}. Jiang \textit{et al.}~\cite{Jiang2018MentorNetLD} proposed MentorNet, which provides a curriculum for a StudentNet to concentrate on the labels that are more likely to be clean. Co-teaching~\cite{Han2018CoteachingRT} trains two networks that teach each other. In particular, they select clean data in the training data, respectively, and feed the clean data to the other network. Other methods take one step further and aim to correct the noisy labels during training to exploit most of the data. Song \textit{et al.}~\cite{Song2019SELFIERU} proposed SELFIE, which corrects refurbishable noisy data for an improvement in the robustness of deep neural networks. DivideMix is another method proposed to correct the noisy labels~\cite{Li2020DivideMixLW}. It first divides the clean and noisy data by modeling their losses and then trains the network on the labeled (clean) data and unlabeled (noisy) data. Since deep neural network models often fit the clean labeled data first~\cite{Wang2021ROBUSTCL}, Liu \textit{et al.}~\cite{liu2020early} proposed a framework called early-learning regularization (ELR), which first produces targets from the model outputs and then regularizes the model towards the targets. In this study, we propose to aggregate and filter noisy labels with external memory to improve the performance for visual sentiment analysis with label noise.

\section{Our Method}
The deep neural networks trained to fit the samples with noisy sentiment labels will generalize poorly on testing cases, and we propose to utilize external memory to aggregate learned knowledge and refine noisy labels. The overview of our method is shown in Figure~\ref{fig:overview}.

Given a training dataset $\lbrace (x_i, y_i) \rbrace_{i=1}^N$ as the input data and the noisy labels from $K$ different classes, a low-dimensional representation $h_i = f_\theta (x_i) \in R^{d}$ is extracted by a network $f$ parameterized by $\theta$. We also have a classifier implemented with a fully connected layer parameterized by $\phi$, and the predication for the $i$-th sample is denoted as $r_i$. Moreover, we denote the memory with $L$ slots as $M=\lbrace C, G \rbrace=\lbrace c_i, g_{i} \rbrace_{i=1}^{L}$, where $K \ll L$, $c_i$ is the representation of memorized prototypes and $g_i \in R^{K}$ is the corresponding soft label.

The primary motivation of our method is to aggregate the learned knowledge into the memory and retrieve a refined label from memory for each sample. We update the prototypes and the corresponding labels to represent a set of samples in a local region of the feature space. The learned prototypes and their labels can be regarded as denoising features and labels for the local regions. They can guide the training process to alleviate the influence of noisy labels. One of the critical parts of our method is updating the memory, which we describe in the following section.

\subsection{Memory Reading for Training with Prototype Labels}
We first describe how to read the memory by supposing that there is a memory composed of the prototypes with soft labels used to store the knowledge of the previous training process. We will describe the memory writing rules later in Section~\ref{subsec:writing}. 

Given the $i$-th training sample, we denote its current prediction as $r_i$ and the latent representation as $h_i$. Its pseudo label $t_i$ is obtained with the memory as
\begin{equation} \label{eq:assign}
\begin{split}
   {t_i} = (1-\lambda)& r_i + \lambda v_i,\\
   v_i=\sum_j^L {p_{i,j}g_j} = &\sum_j^L {\frac{\exp(s(h_i, c_j))}{\sum_l^L \exp(s(h_i, c_l))} g_j}   
\end{split}
\end{equation}
where $s(h_i, c_j)$ is the cosine similarity between the latent representation of $i$-th sample and $j$-th prototype, and $p_{i,j}=\frac{\exp(s(h_i, c_j))}{\sum_l^L \exp(s(h_i, c_l))}$. The pseudo label $t_i$ is a weighted average between the current prediction $r_i$ and the memory-retrieved soft label $v_i$. $\lambda \in [ 0, 1 ]$ controls the trade-off. It is worth noting that  if we set $\lambda=1$ to discard the external memory, this process will be transformed into Bootstrap~\cite{reed2014training} and ELR~\cite{liu2020early}.  
The memory-retrieved label $v_i$ is obtained with the attention addressing mechanism by considering the $i$-th sample representation $h_i$ as the query, the prototype representation $c_j$ as the key, and the prototype label $g_j$ as the value. Briefly, we first conduct the softmax between query $h_i$ and key $c_j$, and then use the obtained address to retrieve the label stored in label memory $v$.  

With the obtained pseudo label $t_i$ and the given label $y_i$, we update our model by minimizing the objective as:
\begin{equation} \label{eq:memoryreadloss}
    \min L_{ce} (r_i, y_i) + \frac{\alpha}{N} \sum_i^N \log (1-r_i^T t_i),
\end{equation}
where $\alpha$ is a hyperparameter and is set as $\alpha =3$.
The first term is a commonly used cross-entropy loss for the current prediction $r_i$ and given label $y_i$. We follow the work of Liu \textit{et al.}~\cite{liu2020early} to use the second term to encourage the prediction $r_i$ to be close to the pseudo label $t_i$.

\subsection{Memory Writing} \label{subsec:writing}
We present how to update the memory $M$ in this section. The memory $M$, composed of the prototypes with the soft labels, plays a crucial role in our method, and a desirable set of prototypes should be representative of local regions. 
We basically rely on deep clustering methods to achieve this goal. We adopt the clustering method proposed by \cite{wu2018unsupervised} to encourage the assignment matrix to approximate target indicator matrix. For detail, given $N$ training samples, we update the prototypes by minimizing 
\begin{equation} \label{eq:adfadf}
    \min_P -\frac{1}{N}\sum_{i}^{N} \sum_j^L q_{i,j} \log p_{i,j}.
\end{equation}
where $p_{i,j}$ is from the assignment matrix and  defined in Eq. (\ref{eq:assign}), and 
$q_{i,j}$ is from the target indicator matrix $Q \in R^{N \times L}$ which is obtained by optimizing the following objective~\cite{asano2019self,caron2020unsupervised}:
\begin{equation} \label{eq:protoupdate}
    \max_{Q} Tr(QC^TH) - \xi \sum_{i}^N \sum_j^K Q_{i,j} \log Q_{i,j},
\end{equation}
where $H=\lbrace h_1, h_2, \dots, h_N \rbrace \in R^{d \times N}$ is the matrix of the latent representation for the training data, and $C=\lbrace c_1, c_2, \dots, c_N \rbrace \in R^{d \times L}$ is the prototype matrix. $\xi$ is a hyperparameter and is set to 0.05 following~\cite{caron2020unsupervised}. Eq. (\ref{eq:protoupdate}) is used to obtain target indicator matrix $Q$ and works similarly as the expectation step of Expectation Maximization while Eq.(\ref{eq:adfadf}) is similar to the maximization step. The $Q$ is optimized to make the trace of the multiplication between reconstructed data representation $QC^T$ and original representation $H$ maximized.
For each minibatch iteration, we first solve Eq. (\ref{eq:protoupdate}) to obtain $Q$ and then encourage $P$ to approximate $Q$ by optimizing Eq. (\ref{eq:adfadf}). We perform end-to-end training through $P$ as it is computed from the latent representation and block the back-propagation through $Q$~\cite{caron2020unsupervised,wu2018unsupervised}. To accommodate Eq. (\ref{eq:protoupdate}) for the mini-batch training, we cache 1K features~\cite{caron2020unsupervised,wu2018unsupervised}. 

In the end, we update the prototype label $g_j \in R^K$ for the $j$-th prototype in a momentum fashion, and $g_j$ is updated to aggregate and memorize the previous predictions of the samples assigned to it as
\begin{equation} \label{eq:protolabelupdate}
    g_j = \beta g_j + (1-\beta) \sum_i q_{i,j}r_i,
\end{equation}
where $\beta$ is the momentum factor and we set $\beta=0.8$. As shown in Eq. (\ref{eq:protolabelupdate}), $g_j$ is updated with the average of the predictions of the samples assigned to it.

\begin{algorithm*}[h]
\DontPrintSemicolon
\SetAlgoLined
\SetNoFillComment
\LinesNotNumbered
 \KwData{the training set: $\mathcal{D} = \{(x_i,y_i)|x\in R^{d_x}, y\in\{0,\cdots,K-1\}\}^{N}_{i=1}$}
 \KwData{memory slot $L$}
 \KwResult{the trained weight $\theta, \phi$}
 \tcc{Model initialization}
 initialize $C$ with an orthogonal matrix;
 
 \tcc{Main loop}
 \While{the maximal iterations are not reached}{
    $x, y \sim \mathcal{D}$ \tcp*{sample training data}
    \tcc{\bf{Prepare for Memory Reading}}
    $v_i \gets \sum_j^L {\frac{\exp(s(h_i, c_j))}{\sum_l^L \exp(s(h_i, c_l))} g_j}$\; \tcp*{retrieve pseudo label from memory}
    ${t_i} \gets (1-\lambda) r_i + \lambda v_i$ \tcp*{generate pseudo label (Eq.~\ref{eq:assign})}
    \tcc{\bf{Prepare for Memory Writing}}
    $Q \gets \arg \max_{Q} Tr(QC^TH) - \xi \sum_{i}^N \sum_j^K Q_{i,j} \log Q_{i,j},$ \tcp*{update the target indicator matrix (Eq.~\ref{eq:protoupdate})}
    \tcc{\bf{Model update}}
    $\theta, \phi , C \gets \arg  \min_{\theta, \phi, C} L_{ce} (r_i, y_i) + \frac{\alpha}{N} \sum_i^N \log (1-r_i^T t_i)  -\frac{1}{N}\sum_{i}^{N} \sum_j^L q_{i,j} \log p_{i,j}$ \tcp*{(Eq.~\ref{eq:overall})}
    \tcc{\bf{Prototype label update}}
    $g_j = \beta g_j + (1-\beta) \sum_i q_{i,j}r_i,$ \tcp*{update the prototype label (Eq.~\ref{eq:protolabelupdate})}

 }
 \caption{Training algorithm.}
 \label{algorithm:training}
\end{algorithm*}


\subsection{Overall Training and Inference}
We summarize the whole training procedure of our method as follows. We initialize the prototype with an orthogonal matrix and update it with backpropogation. For each iteration, we first simultaneously update the network, classifier, and memorized prototypes by minimizing Eq. (\ref{eq:memoryreadloss}) and Eq. (\ref{eq:adfadf}) as 
\begin{equation} \label{eq:overall}
        \min L_{ce} (r_i, y_i) + \frac{\alpha}{N} \sum_i^N \log (1-r_i^T t_i)  -\frac{1}{N}\sum_{i}^{N} \sum_j^L q_{i,j} \log p_{i,j}.
\end{equation}
Second, we update the pseudo labels for prototypes by Eq. (\ref{eq:protolabelupdate}). We use the feature extractor and the classifier to conduct sentiment classification during inference. We summarize the training detail in Alg. (\ref{algorithm:training}).

\begin{figure*}
\centering
\includegraphics[width=1\textwidth]{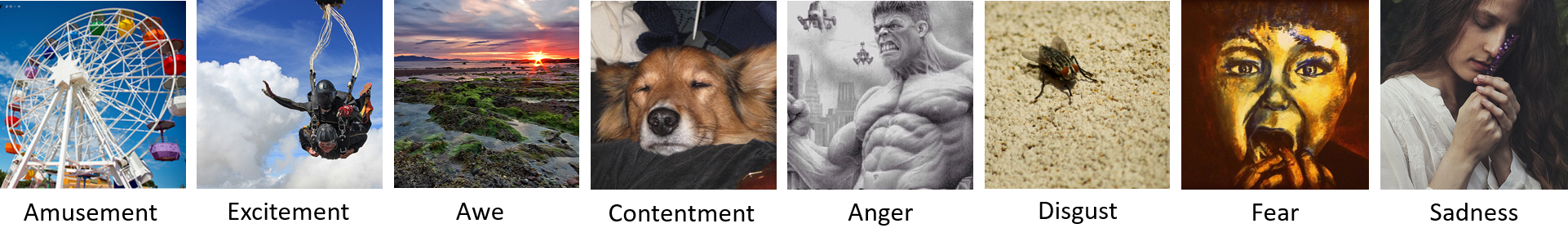}
\caption{Example images from the F\&I dataset.}
\label{fig:example}
\end{figure*}
\section{Experiments}
\subsection{Datasets}
We evaluate our method on publicly available datasets including Flickr and Instagram (F\&I)~\cite{You2016BuildingAL}, Emotion6~\cite{peng2015mixed}, ARTphoto~\cite{machajdik2010affective}, and UnBiasedEmo~\cite{panda2018contemplating}. The F\&I dataset is obtained by querying Flickr and Instagram and contains 21,828 images in total. Emotion6 is retrieved from Flickr with 1,980 images in six sentiment categories. ARTphoto is constructed by 807 images retrieved from an art sharing site, where we convert the task to binary classification (positive/negative). UnBiasedEmo contains about 3000 images from Google in 6 categories. We summarize the statistics of all the datasets in Table \ref{tab:dataset}.

\begin{table}[]
\centering
\caption{Summary of the datasets.}
\begin{tabular}{|c|c|c|}
\hline
Dataset     & Size  & Labeling                                     \\ \hline
F\&I~\cite{You2016BuildingAL}       & 23,308 & \begin{tabular}[c]{@{}c@{}}Awe, amusement, contentment, \\ excitement, disgust, anger, fear, sad\end{tabular} \\ \hline
Emotion6~\cite{peng2015mixed}    & 1,980 & Anger, disgust, fear, joy, sadness, surprise \\ \hline
ArtPhoto~\cite{machajdik2010affective} & 807    & \begin{tabular}[c]{@{}c@{}}Pos, Neg\end{tabular} \\ \hline
UnBiasedEmo~\cite{panda2018contemplating} & 3,045 & Anger, fear, joy, love, sadness, surprise    \\ \hline
\end{tabular}
\label{tab:dataset}
\end{table}

\begin{table*}[]
\centering
\caption{Results on Emotion6 (the best results are highlighted in bold).}
\begin{tabular}{@{}l|cc|cc|cc|cc@{}}
\toprule

\multirow{2}{*}{Methods}     & \multicolumn{2}{c}{SentiBank} & \multicolumn{2}{c}{$\epsilon=0.2$} & \multicolumn{2}{c}{$\epsilon=0.3$} & \multicolumn{2}{c}{$\epsilon=0.4$} \\ 
           & ACC        & F1         & ACC        & F1         & ACC        & F1         & ACC        & F1         \\ \hline
Baseline   & 0.4369     & 0.4102     & 0.4874     & 0.4765     & 0.4548     & 0.4496     & 0.4167     & 0.4114     \\
Bootstrap~\cite{reed2014training}  & 0.4470      & 0.4210      & 0.5000        & 0.4912     & 0.4798     & 0.4695     & 0.4268     & 0.4150      \\
GSE~\cite{zhang2018generalized}        & 0.4621     & 0.4342     & 0.5000        & 0.4851     & 0.4666     & 0.4557     & 0.4313     & 0.4253     \\
SL~\cite{Wang2019SymmetricCE}         & 0.4646     & 0.4359     & 0.4975     & 0.4861     & 0.4722     & 0.4457     & 0.4261     & 0.4146     \\
TCE~\cite{feng2020can}        & 0.4672     & 0.4472     & 0.4848     & 0.4761     & 0.4343     & 0.4252     & 0.4596     & 0.4564     \\
CT+~\cite{yu2019does} & 0.4343     & 0.4143     & 0.5051  & 0.4899  & 0.4922  & 0.4793  & 0.4646  & 0.4584 \\
MixUp~\cite{zhang2017mixup}      & 0.4619     & 0.4327     & 0.5278     & \textbf{0.5172}     & 0.4672     & 0.4624     & 0.4545     & 0.4479     \\
ELR~\cite{liu2020early}        & 0.4571     & 0.4421     & 0.4949     & 0.4775     & 0.4848     & 0.4634     & 0.4192     & 0.4127     \\ \hline
Ours       &  \textbf{0.4747}          & \textbf{0.4486}            & \textbf{0.5303}     & 0.5144     & \textbf{0.5278}     &  \textbf{0.5127}     & \textbf{0.4823}     & \textbf{0.4714}     \\ \bottomrule
\end{tabular}
\label{tab:resultsmislabelEmotion6}
\end{table*}

\begin{table*}[]
\centering
\caption{Results on ArtPhoto (the best results are highlighted in bold).}
\begin{tabular}{@{}l|cc|cc|cc|cc@{}}
\toprule
\multirow{2}{*}{Methods}     & \multicolumn{2}{c}{SentiBank} & \multicolumn{2}{c}{$\epsilon=0.2$} & \multicolumn{2}{c}{$\epsilon=0.3$} & \multicolumn{2}{c}{$\epsilon=0.4$} \\ 
           & ACC        & F1         & ACC        & F1         & ACC        & F1         & ACC        & F1         \\ \hline
Baseline   & 0.6181                        & 0.6151                        & 0.6975                  & 0.6966                 & 0.679                   & 0.6766                 & 0.6235                  & 0.6233                 \\
Bootstrap~\cite{reed2014training}  & 0.6220                         & 0.6194                        & 0.7022                  & 0.6981                 & 0.6358                  & 0.6355                 & 0.6111                  & 0.6077                 \\
GSE~\cite{zhang2018generalized}        & 0.6171                        & 0.6114                        & 0.6667                  & 0.6667                 & 0.6728                  & 0.6728                 & 0.6605                  & 0.6599                 \\
SL~\cite{Wang2019SymmetricCE}         & 0.6296                        & 0.6294                        & 0.6975                  & 0.6966                 & 0.6667                  & 0.6592                 & 0.6420                   & 0.6406                 \\
TCE~\cite{feng2020can}        & 0.6235                        & 0.6110                         & 0.7184                  & 0.7077                 & 0.6975                  & 0.6969                 & 0.6852                  & 0.6798                 \\
CT+~\cite{yu2019does} & 0.6543                        & 0.6530                         &0.7160     & \textbf{0.7133}    & 0.6667    & 0.6648   & 0.6296    &0.6291   \\
MixUp~\cite{zhang2017mixup}      & 0.5988                        & 0.5953                        & 0.7043                  & 0.7018                 & 0.6958                  & 0.6925                 & 0.6914                  & 0.6913                 \\
ELR~\cite{liu2020early}        & 0.6296                        & 0.6287                        & 0.6852                  & 0.6798                 & 0.6728                  & 0.6692                 & 0.6605                  & 0.6594                 \\ \hline
Ours       & \textbf{0.6675} & \textbf{0.6581} & \textbf{0.7222}                  & 0.7117                 & \textbf{0.7099}                  & \textbf{0.7080}                  & \textbf{0.7037}                  & \textbf{0.7015}                 \\ \bottomrule
\end{tabular}
\label{tab:resultsmislabelArt}
\end{table*}
\subsection{Experimental Settings and Evaluation Metrics}
We run an ordinal cross-entropy classifier as the baseline, and additionally compare the proposed ARNet with other learning-with-noisy-label methods including Bootstrap~\cite{reed2014training}, GSE~\cite{zhang2018generalized}, SL~\cite{Wang2019SymmetricCE}, TCE~\cite{feng2020can}, CoTeaching+(CT+)~\cite{yu2019does}, MixUp~\cite{zhang2017mixup}, and ELR~\cite{liu2020early}. For ARNet, we set $\lambda=0.8$, $\beta=0.8$, and search the number of memory slots from $\lbrace 16, 32, 64, 128 \rbrace$. 
\begin{table}[]
\centering
\caption{Results on UnBiasedEmo (the best results are highlighted in bold).}
\begin{tabular}{@{}l|cc|cc|cccc@{}}
\toprule
\multirow{2}{*}{Methods}     & \multicolumn{2}{c}{$\epsilon=0.2$} & \multicolumn{2}{c}{$\epsilon=0.3$} & \multicolumn{2}{c}{$\epsilon=0.4$} \\ 
                & ACC        & F1         & ACC        & F1         & ACC        & F1         \\ \hline
Baseline   & 0.5419 & 0.5167 & 0.4614 & 0.4313 & 0.4072 & 0.3678 \\
Bootstrap~\cite{reed2014training}  & 0.5238 & 0.4937 & 0.4614 & 0.4321 & 0.4171 & 0.3871 \\
GSE~\cite{zhang2018generalized}        & 0.5543 & 0.4800   & 0.5189 & \textbf{0.4934} & 0.4729 & 0.4000    \\
SL~\cite{Wang2019SymmetricCE}         & 0.5222 & 0.4992 & 0.5140  & 0.4800   & 0.4417 & 0.3978 \\
TCE~\cite{feng2020can}          & \textbf{0.5649} & \textbf{0.5325} & 0.5189 & 0.4659 & 0.4154 & 0.3589 \\
CT+~\cite{yu2019does} & 0.5454       &0.4568        &0.4696  &0.4156      &  0.4663       & 0.3940       \\
MixUp~\cite{zhang2017mixup}      & 0.5419 & 0.5106 & 0.4778 & 0.4472 & 0.4401 & 0.4041 \\
ELR~\cite{liu2020early}        & 0.5419 & 0.5074 & 0.4729 & 0.4495 & 0.4548 & 0.3934 \\ \hline
Ours       & 0.5517 & 0.5184 & \textbf{0.5238} & 0.4867 & \textbf{0.4744} & \textbf{0.4108} \\ \bottomrule 
\end{tabular}
\label{tab:resultsmislabelUnbiasedEmo}
\end{table}
\begin{table}[]
\centering
\caption{Results on F1 (the best results are highlighted in bold).}
\begin{tabular}{@{}l|cc|cc|cc@{}}
\toprule
\multirow{2}{*}{Methods}     & \multicolumn{2}{c}{$\epsilon=0.2$} & \multicolumn{2}{c}{$\epsilon=0.3$} & \multicolumn{2}{c}{$\epsilon=0.4$} \\
           & ACC        & F1         & ACC        & F1         & ACC        & F1         \\ \hline
Baseline   & 0.5197     & 0.4722     & 0.4986     & 0.4207     & 0.4528     & 0.3977     \\
Bootstrap~\cite{reed2014training}  & 0.5588     & 0.5026     & 0.5014     & 0.4419     & 0.4647     & 0.3967     \\
GSE~\cite{zhang2018generalized}        & 0.5765     & 0.4285     & 0.5619     & 0.4239     & 0.5307     & 0.3957     \\
SL~\cite{Wang2019SymmetricCE}         & 0.5655     & \textbf{0.5117}     & 0.5032     & 0.4527     & 0.5038     & 0.4372     \\
TCE~\cite{feng2020can}        & 0.5850      & 0.4518     & 0.5383     & 0.4385     & 0.4913     & 0.4337     \\
CT+~\cite{yu2019does} &0.5481            &0.4544    &0.5307       & 0.4066     &  0.4922      &0.3862            \\
MixUp~\cite{zhang2017mixup}      & 0.5527     & 0.4860      & 0.5111     & 0.4597     & 0.4782     & 0.4140      \\
ELR~\cite{liu2020early}        & 0.5481     & 0.4833     & 0.5160      & 0.4449     & 0.4565     & 0.3863     \\ \hline
Ours       & \textbf{0.5887}     & 0.5026     & \textbf{0.5655}     & \textbf{0.4951}     & \textbf{0.5337}     & \textbf{0.4668}     \\ \bottomrule
\end{tabular}
\label{tab:resultsmislabelFI}
\end{table}

For Emotion6, ARTphoto, and Unbiased Emotion, we randomly split the dataset into 80\%  training set and 20\%  testing set following previous work~\cite{She2020WSCNetWS,xue2020nlwsnet}. For F\&I, we split the data into 80\% training set, 5\% validation set and 15\% testing set following~\cite{She2020WSCNetWS}. We adopt an ImageNet pre-trained ResNet-34 as the feature extractor~\cite{he2016deep}, and use the Adam optimizer with a learning rate of 0.0001. The batchsize is set to 128 and we run all methods for 50 epoches for these datasets. We first resize the images to 256x256, and then perform random crop and random horizontal flip for data augmentation to obtain images of 224*224. We report accuracy and F1 scores on the testing set averaged over three runs. We use PyTorch to implement ARNet, and all the experiments are run on a Linux server with 4x2080Ti Graphical Cards.

We propose two different settings to simulate the mislabeled scenarios for visual sentiment analysis. For the first scenario, we follow the conventional settings for learning with noisy labels~\cite{Han2018CoteachingRT, feng2020can,liu2020early,zhou2021robust}, and symmetrically flip labels for $\epsilon$ \% of training samples. We vary $\epsilon \in \lbrace 0.2, 0.3, 0.4 \rbrace$. We simulate a virtual agent with a weak classifier and use the classification prediction as ``pseudo"  noisy labels for the second scenario. We expressly adopt SentiBank with an SVM classifier as the virtual agent. We train on the noisy label set for both settings and test on the original dataset.   

\subsection{Experimental Results}
The experimental results are shown in Tables \ref{tab:resultsmislabelEmotion6}, \ref{tab:resultsmislabelArt}, \ref{tab:resultsmislabelUnbiasedEmo}, and \ref{tab:resultsmislabelFI}. We draw several conclusions as follows. 
First, the results show that deep models in learning with noisy labels can boost visual sentiment analysis performance with label noise. For instance, our method achieves 5.3\% improvements in terms of accuracy compared with the baseline on Emotion6 on average, and 4.7\% on ArtPhoto. The encouraging results suggest that it is possible to train a robust classifier for visual sentiment analysis with webly retrieved data. Moreover, we note that the proposed visual sentiment benchmark can also work as a benchmark suite to evaluate methods of learning with noisy labels.
\begin{figure*}
\centering
\subfigure[Baseline]{
\includegraphics[width=0.31\textwidth]{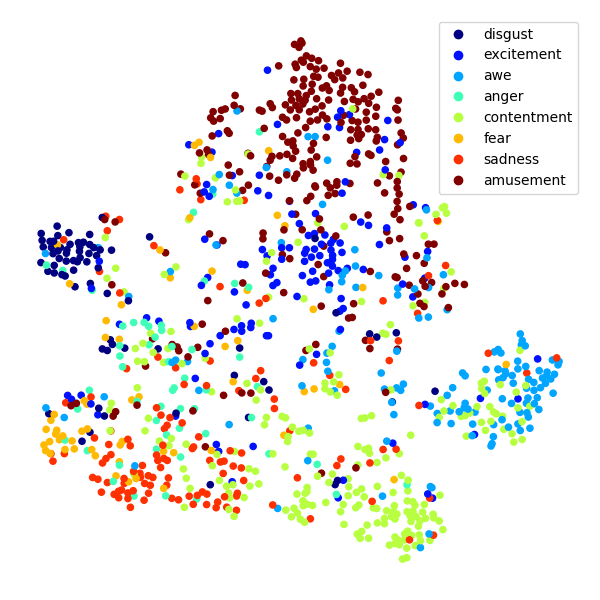}
}
\subfigure[ELR]{
\includegraphics[width=0.31\textwidth]{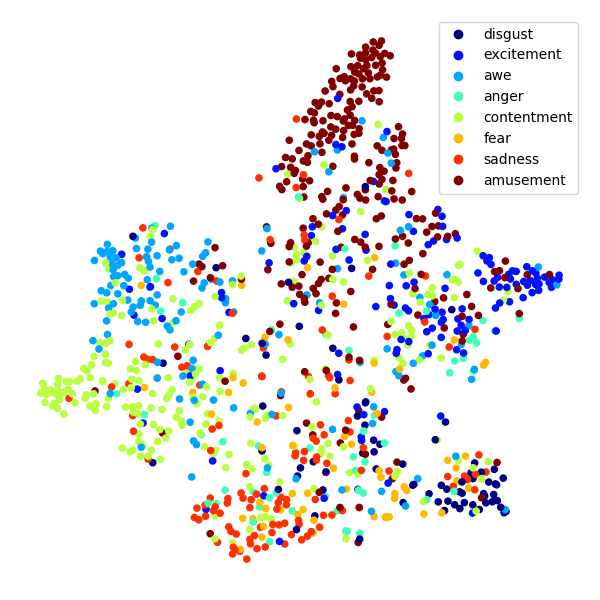}
}
\subfigure[Ours]{
\includegraphics[width=0.31\textwidth]{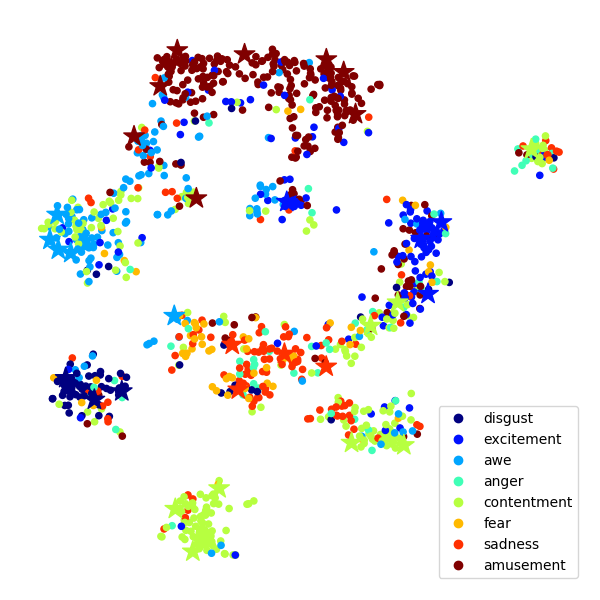}
}
\caption{TSNE visualization of the learned embedding for the baseline, ELR, and our method on the F\&I validation set. Circles denote latent representations of the samples, and stars denote memorized prototypes.}
\label{fig:vis}
\end{figure*}

\begin{table*}[!htb]
    \begin{minipage}{.5\linewidth}
      \centering
      \caption{Results on F\&I with different number of memory slots.}
      \small
        \begin{tabular}{@{}lccccc@{}}
        \toprule
        {$L$} & 16 & 32     & 64     & 128 \\ \midrule
        ACC       &0.5195   &0.5314 & 0.5655 &0.5714     \\ \midrule
        F1       &0.4512   & 0.4683 & 0.4951 & 0.4983   \\ \bottomrule
        \end{tabular}
        \label{tab:multishot}
    \end{minipage}%
    \begin{minipage}{.5\linewidth}
      \centering
      \caption{Results on UnBiasedEmo with different number of memory slots.}
      \small
        \begin{tabular}{@{}lccccc@{}}
        \toprule
        {$L$}  & 16  &32   & 64      & 128\\ \midrule
        ACC       & 0.4762   & 0.4975 & 0.5238    & 0.4992   \\ \midrule
        F1       &  0.4400  &0.4511  &    0.4867 & 0.4656 \\ \bottomrule
        \end{tabular}
        \label{tab:multiway}
    \end{minipage} 
\end{table*}
Second, the proposed ARNet consistently outperforms the state-of-the-art methods for handling noisy sentiment labels, for example,  about 2.7\%
improvement on F\&I and 2.1\% on ArtPhoto in terms of average accuracy, which demonstrates
ARNet can learn more discriminative and informative representation with the proposed external memory. The memory also allows the training samples to share and mutually refine their noisy labels. We validate this point by visualizing the latent representation obtained by different models in Fig.~\ref{fig:vis}, where the visualization results directly show that the proposed ARNet could learn to aggregate and refine noisy labels via the external memory and the locality representative prototypes. Fig. \ref{fig:vis} also suggests that the learned representations of ARNet are optimized to have small intra-class distance and large inter-class distance, which should be attributed to the fact that we jointly conduct prototype optimization and noisy label refining in Eq. (\ref{eq:overall}). The excellent property of the learned representation directly leads to better performance by ARNet compared with other methods.

Finally, by comparing two different scenarios for visual sentiment analysis with label noise,  the proposed ``virtual" agent by the SentiBank sentiment classifier brings more challenges to existing methods as the mislabeled samples by the ``virtual" agent is more challenging and more confusing. Our method also achieves superior performance in the virtual agent setting. For example, the proposed ARNet achieves more than 1.3\% and 0.5\% performance gain in terms of accuracy and F1 score, respectively, on the ArtPhoto dataset. 

%

\subsection{Ablation Study}
In this section, we conduct experiments to study the influence of the hyper-parameters on F\&I and UnBiasedEmo with $\epsilon=0.3$. We vary the number of memory slots from $\lbrace 16, 32, 64, 128 \rbrace$, and the results are shown in Table \ref{tab:multishot} and Table \ref{tab:multiway}, respectively. The results conclude that better performance can be expected with an increasing number of memory slots, and the incremental improvements will be negligible when the memory slots are large enough. For example, for UnBiasedEmo, it seems that 64 prototype slots are adequate, and 128 prototype lots may lead to an overfitting problem. 




\section{Conclusions}
This paper presents ARNet to handle label noise in visual sentiment analysis. ARNet utilizes an external memory to aggregate and filter the noisy labels and provide refined labels for training samples. Moreover, we propose reading the memory through attention addressing mechanism and updating the memory to have the stored prototypes be locality representative. The prototype labels can be used to aggregate and refine the labels for samples in the local region of the feature space. The learned prototypes and labels can be regarded as denoising features and labels for the local regions. We establish a benchmark for visual sentiment analysis with label noise using publicly available datasets, and our experiments show the effectiveness of our method the proposed benchmark.  
\bibliographystyle{IEEEtran}
\bibliography{egbig}
\end{document}